\documentclass[runningheads]{llncs}
\usepackage{graphicx}
\usepackage[numbers]{natbib}
\usepackage[symbol]{footmisc}

\usepackage{amsmath}
\usepackage{amssymb}
\usepackage{booktabs}
\usepackage{array}
\usepackage{changepage}
\usepackage[pagebackref,breaklinks,colorlinks]{hyperref}
\usepackage[capitalize]{cleveref}
\crefname{section}{Sec.}{Secs.}
\Crefname{section}{Section}{Sections}
\Crefname{table}{Table}{Tables}
\crefname{table}{Tab.}{Tabs.}
\newcolumntype{X}[1]{>{\centering\let\newline\\\arraybackslash\hspace{0pt}}m{#1}}
\newcolumntype{Y}[1]{>{\raggedright\let\newline\\\arraybackslash\hspace{0pt}}m{#1}}
\begin{document}

\title{\large A Graphical Approach to Document Layout Analysis}

\author{Jilin Wang\inst{1,*} \and Michael Krumdick\inst{1,*} \and Baojia Tong\inst{2} \and Hamima Halim\inst{1} \and Maxim Sokolov\inst{1} \and Vadym Barda\inst{1} \and Delphine Vendryes\inst{3} \and Chris Tanner\inst{1}}

\authorrunning{J. Wang \& M. Krumdick et al.}
%
\institute{Kensho Technologies, Cambridge, MA, 02138, USA\\
\email{\{jilin.wang, michael.krumdick, hamima.halim, maxim.sokolov, \\vadym, chris.tanner\}@kensho.com} \and
Meta, Cambridge, MA, 02140, USA \\
\email{tongbaojia@gmail.com}  \and
Google, Los Angeles, CA, 90291, USA \\
\email{delphine.vendryes@gmail.com}}

\maketitle
\footnotetext[1]{The authors contributed equally to this work.}
\begin{abstract}

Document layout analysis (DLA) is the task of detecting the distinct, semantic content within a document and correctly classifying these items into an appropriate category (e.g., text, title, figure). DLA pipelines enable users to convert documents into structured machine-readable formats that can then be used for many useful downstream tasks. Most existing state-of-the-art (SOTA) DLA models represent documents as images, discarding the rich metadata available in electronically generated PDFs. Directly leveraging this metadata, we represent each PDF page as a structured graph and frame the DLA problem as a graph segmentation and classification problem. We introduce the Graph-based Layout Analysis Model (GLAM), a lightweight graph neural network competitive with SOTA models on two challenging DLA datasets - while being an order of magnitude smaller than existing models. In particular, the 4-million parameter GLAM model outperforms the leading 140M+ parameter computer vision-based model on 5 of the 11 classes on the DocLayNet dataset. A simple ensemble of these two models achieves a new state-of-the-art on DocLayNet, increasing mAP from 76.8 to 80.8. Overall, GLAM is over 5 times more efficient than SOTA models, making GLAM a favorable engineering choice for DLA tasks. 

\end{abstract}


\section{Introduction}
\label{sec:intro}

Much of the world's information is stored in PDF (Portable Document Format) documents; according to data collected by the Common Crawl\footnote{https://commoncrawl.github.io/cc-crawl-statistics/plots/mimetypes}, PDF documents are the most-popular media type on the web other than HTML.

A standard PDF file contains a complete specification of the rendering of the document. All objects (e.g., texts, lines, figures) in the PDF file include their associated visual information, such as font information and location on the page. However, a PDF file does not always contain information about the \textit{relationships} between those objects, which is critical for understanding the structure of the document. For example, all text is stored at the individual character level. Sentences from the characters and consequently paragraphs or titles formed by the sentences are not included in the metadata. Therefore, despite PDF files containing explicit, structured representations of their content, correctly structuring this content into human-interpretable categories remains a challenging problem and is the crux of Document Layout Analysis (DLA).

\begin{figure*}
\begin{adjustwidth}{-2.8cm}{}
     \centering
     \includegraphics[width=1.2\linewidth]{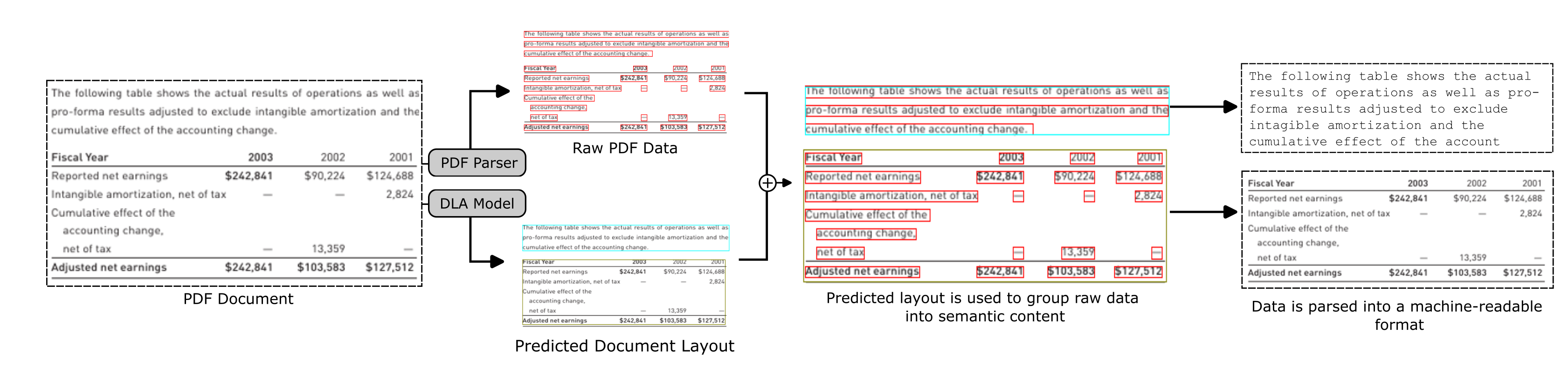}
\end{adjustwidth}
     \caption{Visualization of the data extraction pipeline. We use {\it pdfminer.six} as the PDF parser in our pipeline.}
     \label{fig:dla_pipeline}
\end{figure*}

Existing approaches to PDF extraction typically consist of two main components --- a PDF parser and DLA model --- as depicted in \cref{fig:dla_pipeline} \cite{8892913,10.1007/978-3-540-28640-0_20}:
\begin{enumerate}
    \item \textbf{PDF parser} (e.g., \textit{PDF Miner}): extracts the document's underlying raw data, often doing so by producing fine-grained, pertinent contiguous blocks of information referred to as ``objects'' (e.g., a line of text). For a standard (non-scanned) PDF document, the PDF parser is able to generate information about each ``object'': its location on the page, the raw content contained within it, and some metadata like font and color about how the object is displayed.
    \item \textbf{DLA model:} aims to group together objects that concern semantically similar content (e.g., a table, figure, paragraph of text), essentially turning an unordered collection of data into a coherent, structured document. The output is a collection of bounding boxes, each representing one of the semantic objects (e.g., a paragraph).
\end{enumerate}

DLA models frequently take a computer vision-based approach \cite{huang2022layoutlmv3,li2022dit,xu-etal-2021-layoutlmv2}. These approaches rely on treating documents as images and ignoring much of the rich, underlying metadata.

Generally, PDF extraction can be evaluated by the quality of the DLA model's segmentation predictions. The standard approach to evaluating the segmentation performance is to use the mean average precision (mAP) metric to assess the overlap between the predicted and ground-truth bounding boxes. Although this is standard practice, mAP is sensitive to the bounding boxes' exact pixel values. Therefore, bounding box-level metrics are only proxies for measuring segmentation performance, and ultimately a user wishes to measure the performance of clustering objects to semantic segments. In this work, we approach the problem by modeling the document not as a collection of pixels but as a structured graphical representation of the output from a PDF parser. 

Specifically, we frame the DLA task as a joint graph segmentation and node classification problem. Using this formulation, we develop and evaluate a graph neural network (GNN) based model -- the Graph-based Layout Analysis Model (GLAM). Our model is able to achieve performance that is comparable in mAP to the current SOTA methods while being lightweight, easier to train, and faster at inference time.

Our primary contributions are :
\begin{enumerate}
\item Propose a novel formulation of DLA, whereby we create a structured graphical representation of the parsed PDF data and treat it as a joint node classification problem and graph segmentation problem.
\item Develop a novel Graph-based Layout Analysis Model (GLAM) for this problem formulation and evaluate it on two challenging DLA tasks: PubLayNet and DocLayNet.
\item Demonstrate that by ensembling our model with an object detection model, we are able to set a new state-of-the-art (SOTA) on DocLayNet, increasing the mAP on this task from 76.8 to 80.8.  
\item Validate that GLAM is compact in size and fast at speed with a comparison against SOTA models, making it efficient and applicable in industrial use cases.
\end{enumerate}

\section{Related Work}
\subsection{Document Layout Analysis}
As a field, document layout analysis (DLA) is mainly aimed toward enterprise use cases \cite{binmakhashen2019document}. Businesses often need to process massive streams of documents to complete complex downstream tasks, such as information retrieval, table extraction, key-value extraction, etc. Early literature in this space focused on parsing-based solutions that would attempt to understand the document layout from bounding boxes and textual data parsed from born-digital files, leveraging features from the native file type \cite{10.1007/978-3-540-28640-0_20}. 

\subsection{Object Detection-Based Methods in VRDU}
Recent work has focused on Visually Rich Document Understanding (VRDU), which models documents as images. Image-based object detection formulations of this problem have been particularly favorable and popular because of their generality; effectively any type of human-readable document can be converted into an image. Further, using a traditional, simple convolutional neural network (CNN) for feature extraction works well in generating representations of documents. Early iterations of \textit{LayoutLM} and others \cite{xu2020layoutlm,gu2022unified} all adopt this method. 

Optionally, textual information can be captured by an optical character recognition (OCR) engine, avoiding the need to integrate text parsing engines for multiple document media types \cite{cui2021document}. This OCR step, whose accuracy is critical for any model that uses textual information \cite{li-etal-2021-structurallm}, could be fine-tuned on specific formats but will also introduce an upper bound of OCR accuracy. 

Recently, there has been an emphasis on using large pre-trained document understanding models that leverage vast, unlabeled corpora, such as the IIT-CDIP test collection \cite{10.1145/1148170.1148307}. The Document Image Transformer (DiT) \cite{li2022dit} model adopts a BERT-like pre-training step to train a visual backbone, which is further adapted to downstream tasks, including DLA. There have been additional efforts in creating pre-trained multi-modal models that leverage both the underlying document data and the visual representation. The LayoutLM series of models \cite{xu-etal-2021-layoutlmv2,huang2022layoutlmv} have been particularly successful. LayoutLMv3 \cite{huang2022layoutlmv3} is pre-trained with Masked Language Modeling (MLM), Masked Image Modeling (MIM), and Word-Patch Alignment (WPA) training objectives. The LayoutLMv3 is integrated as a feature backbone in a Cascade R-CNN detector \cite{cai2018cascade} with FPN \cite{lin2017feature} for downstream DLA tasks.

\subsection{Graph Neural Networks in Document Understanding}

Graphs and graphical representations have been used extensively in document understanding (DU) \cite{10.1016/j.patcog.2010.11.015}, especially toward the problem of table understanding. \cite{10.1007/978-3-030-30645-8_27} extracted saliency maps using a CNN and then used a Conditional Random Field (CRF) model to reason about the relationships of the identified salient regions for table and chart detection. \citet{8892877} used the parsed data from a PDF to create a graphical representation of a document and then a GNN for table detection and extraction. \citet{8978070} uses a computer-vision feature augmented graph and a GNN for segmentation in order to perform table structure recognition. 

Graphs have been used in DLA as well. \citet{10.1007/978-3-030-86549-8_8} use data from a PDF parser and a module based on graph neural networks to model the relationship of potential candidate layout outputs as part of their larger, multi-modal fusion architecture. A key component of any graph-based method is the formulation of the graphs. While the edges have diverse interpretations, the nodes will typically represent some logically contiguous region of the document. In image-based methods, this can be obtained by computer vision, as in \cite{li_gat_2017}, or by a more modular OCR task as leveraged by \cite{Liu2019GraphCF}. In contrast, \citet{graphNLP_wei_2020} and this work parse the ground truth text boxes if focusing on digital-generated documents.

Typically, two nodes are connected by an edge if they satisfy certain spatial criteria (e.g., based on their vertical or horizontal locations). 
\citet{ribapaper} notes that the resulting graph structures are similar to visibility graphs, which represent nodes as geometric blocks of space that are connected only if they have some vertical or horizontal overlap with no other nodes in between.

Most similar to our work are \cite{li_gat_2017} and \cite{9956590}. \citet{li_gat_2017} also frames DLA as a graph segmentation and node classification problem but uses computer vision in order to build its ``primitive'' nodes rather than using PDF parsing. \citet{9956590} uses a GNN on a representation of the document similar to this work. However, they focus only on node classification for table extraction. Our model is able to perform the entire DLA task in an end-to-end fashion by modeling both node-level classifications and graph segmentation.

\section{Methodology}
\subsection{Graph Generation}

\begin{figure}[h]
     \centering
     \includegraphics[width=0.8\textwidth]{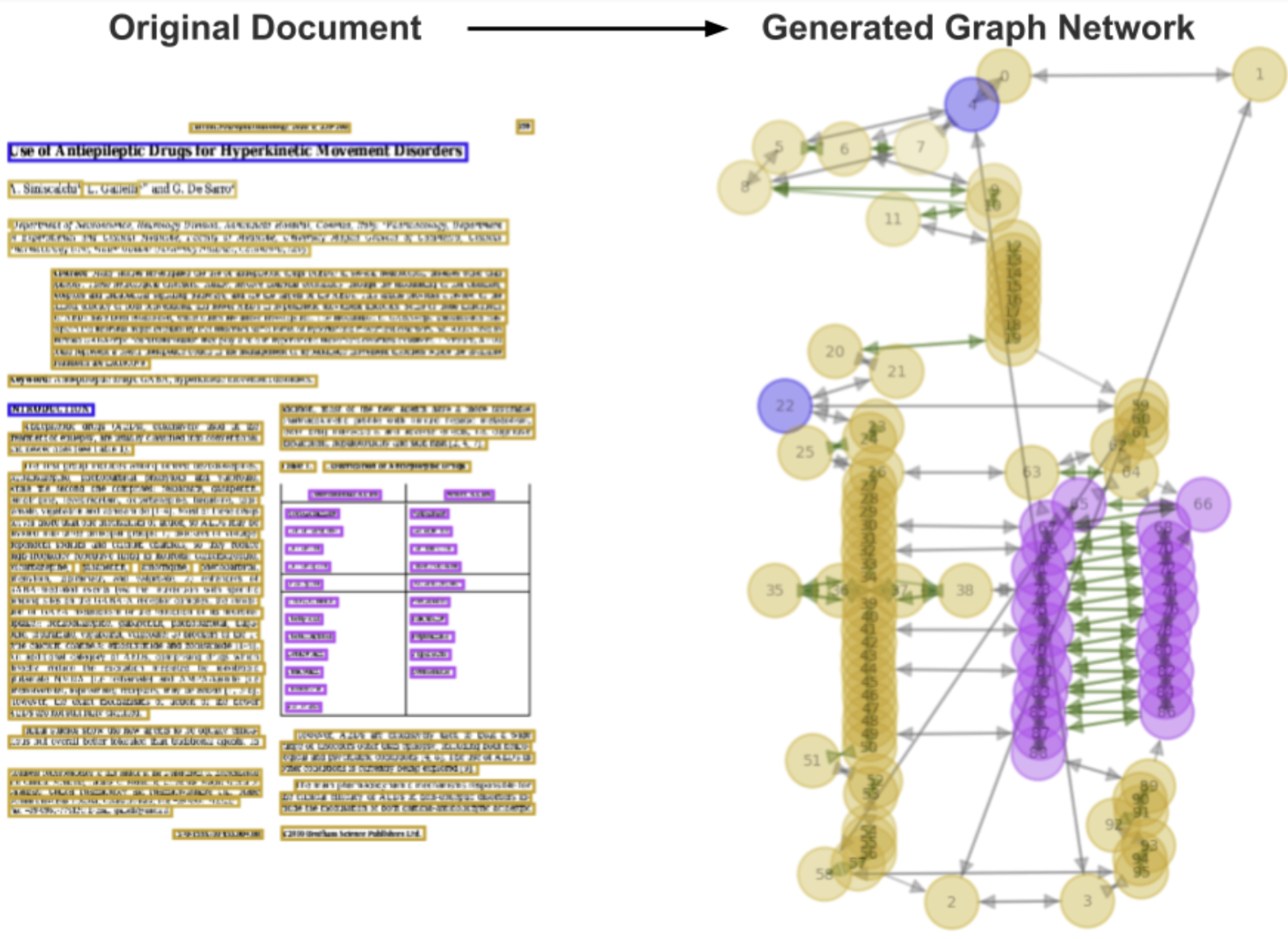}
     \caption{Example document graph from PubLayNet with associated node and edge labels. Left: text boxes parsed from the PDF, whereby categories are labeled in different colors (blue for \textit{title}, orange for \textit{text}, and magenta for \textit{table}). Right: corresponding graph generated. Each node represents a text box, and the arrows between nodes represent the edges. Illustratively, green edges indicate a link (positive), and grey edges represent there is no link (negative). }
     \label{fig:graph_example}
\end{figure}

We use a PDF parser to extract all text boxes, each of which also includes 79 associated features, such as bounding box location, text length, the fraction of numerical characters, font type, font size, and other simple heuristics. Using this parsed metadata, we construct a vector representation for each individual box, which constitutes a node in our graph.

Bidirectional edges are established between nodes by finding each node's nearest neighbor in each possible direction (up, down, left, and right). Additional edges are added to approximate the potential reading order of the underlying text boxes in the document (i.e. the heuristic order parsed from PDF documents -- up to down, left to right). Each edge also has its own vector representation, which includes the edge direction and the distance between the two nodes that it connects. A visualization of the resulting graph can be seen in \cref{fig:graph_example}.

\subsection{Graph-based Layout Analysis Model}

There are two key parts to the Graph-based Layout Analysis Model (GLAM): a graph network and a computer vision feature extractor. 

\subsubsection{Graph Network}

\begin{figure}[h]
  \centering
    \includegraphics[width=2 in]{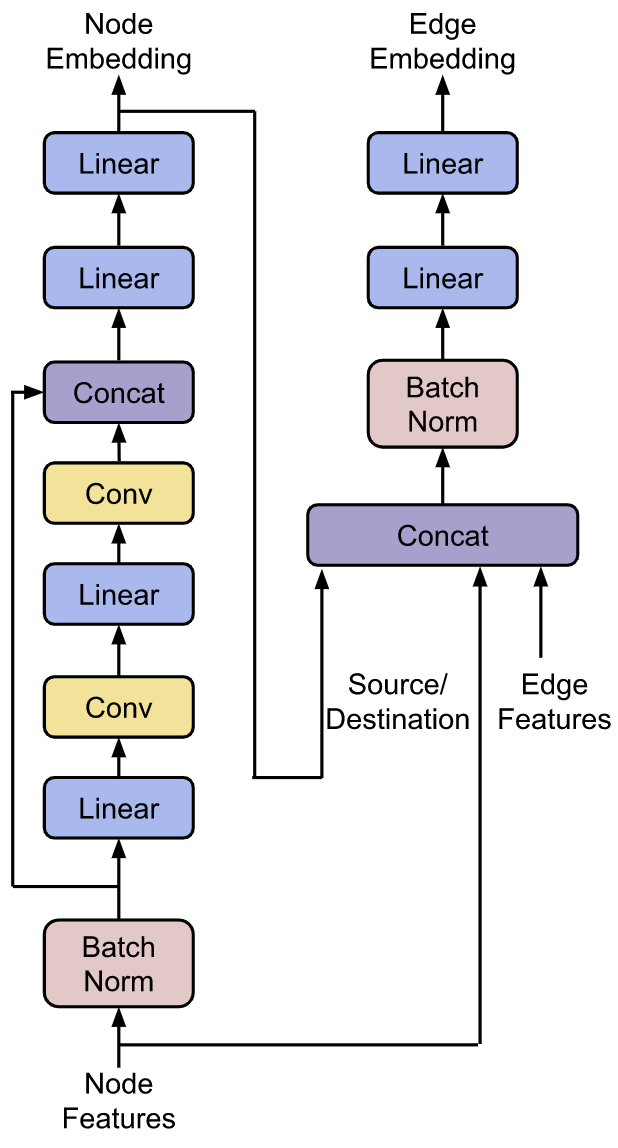}
    \caption{Visualization of the GNN architecture used in GLAM.}
    \label{fig:gnn-model}
\end{figure}
Our graph network model consists of a graph convolutional network (GCN) \cite{DBLP:conf/iclr/KipfW17} and two classification heads -- one for node classification and one for edge classification. The model structure is shown in \cref{fig:gnn-model}. We apply batch normalization to the node features before inputting them into a series of linear and topology adaptive graph (TAG) convolutional \cite{du2017topology} layers. We set the TAG's first hidden layer to have 512 dimensions when training on PubLayNet, and we increase this to 1024 dimensions for the more complex DocLayNet dataset. As the GCN network gets deeper, we scale down the number of parameters in both linear and TAG convolutional layers by a factor of 2. We concatenate this output with the original normalized features and then apply a series of linear layers to create our node embeddings. A classification head then projects this into a distribution over the possible segment classes (e.g., title, figure, text). This model is very lightweight, totaling just over one million parameters.

To create the edge embeddings, we extract the relevant node embeddings and concatenate them with the original edge features. We batch-normalize this representation and then apply another two linear layers to generate the final edge embeddings.  

The model is trained with a weighted, joint loss function \cref{loss} from the two classification tasks.
\begin{equation}
    L = L_{node} + \alpha L_{edge}
\label{loss}
\end{equation}

\noindent where $L_{node}$ and $L_{edge}$ are a cross-entropy loss for node and edge classification, and the edge loss scale $\alpha=4$ is applied to improve edge classification results for better segmentation. 

\begin{figure}[h]
    \centering
    \includegraphics[width=\linewidth]{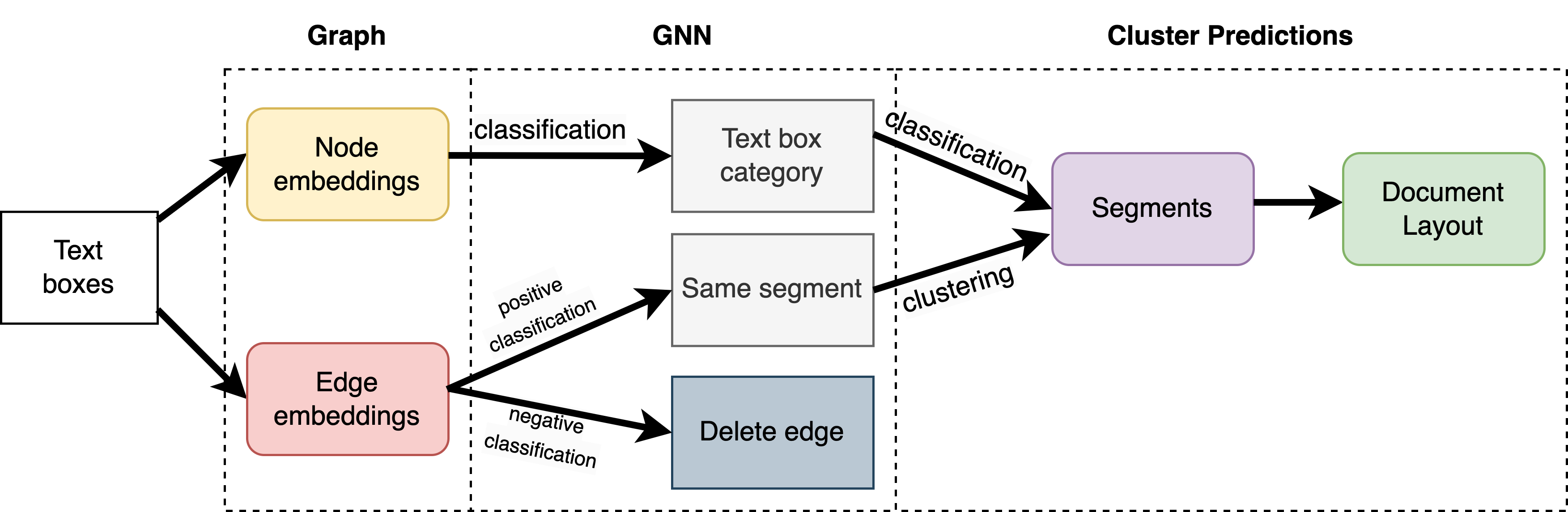}
    \caption{Visualization of the model pipeline.}
    \label{fig:gnn}
\end{figure}

At inference time, segment clustering follows the steps shown in \cref{fig:gnn}. For a given input graph, the model will classify each node into one of the given segment categories and predict whether each edge is positive or negative. The negative edges are removed from the graph. The segments are then defined by the remaining connected components in the graph. Each segment is classified by the majority vote of the classification of all of its component nodes. This is finally converted back into a COCO-style annotation by using the minimum spanning box of the nodes within the segment as the bounding box, the selected class label as the label, and the mean probability of that label over all of the nodes as the score.

\subsubsection{Computer Vision Feature Extraction}
\label{sec:cv}
One of the main limitations of PDF-parser-based data pre-processing is that it is unable to capture all of the visual information encoded in the PDF. All of the text boxes are only represented by their metadata, which does not account for many of the semantic visual cues within a layout. These cues include background colors, lines, dots, and other non-textual document elements. To include this visual information, we extend our node-level features with a set of visual features. We render the PDF as an image and extract a visual feature map using ResNet-18 \cite{7780459}. Specifically, to extract the image features corresponding to each input node's bounding box, we use Region of Interest (ROI) extraction with average pooling \cite{8237584}. 

We concatenate these image features with our metadata-based feature vectors, then apply a simple three-layer attention-based encoder model \cite{NIPS2017_3f5ee243} to these features to give them global context before concatenating them with the features associated with the nodes in our graph. This maintains the overall graphical structure while adding the ability to leverage additional visual cues. The network is also relatively small, totaling only slightly over three million parameters. This feature extraction process does not create any additional nodes. Thus, while the model may be able to include some information from image-rich regions of the document --- such as figures and pictures --- the model is still limited in its ability to account for them.

\section{Experiments}

\subsection{Datasets}

\begin{figure}[h]
     \centering
     \includegraphics[width=0.7\textwidth]{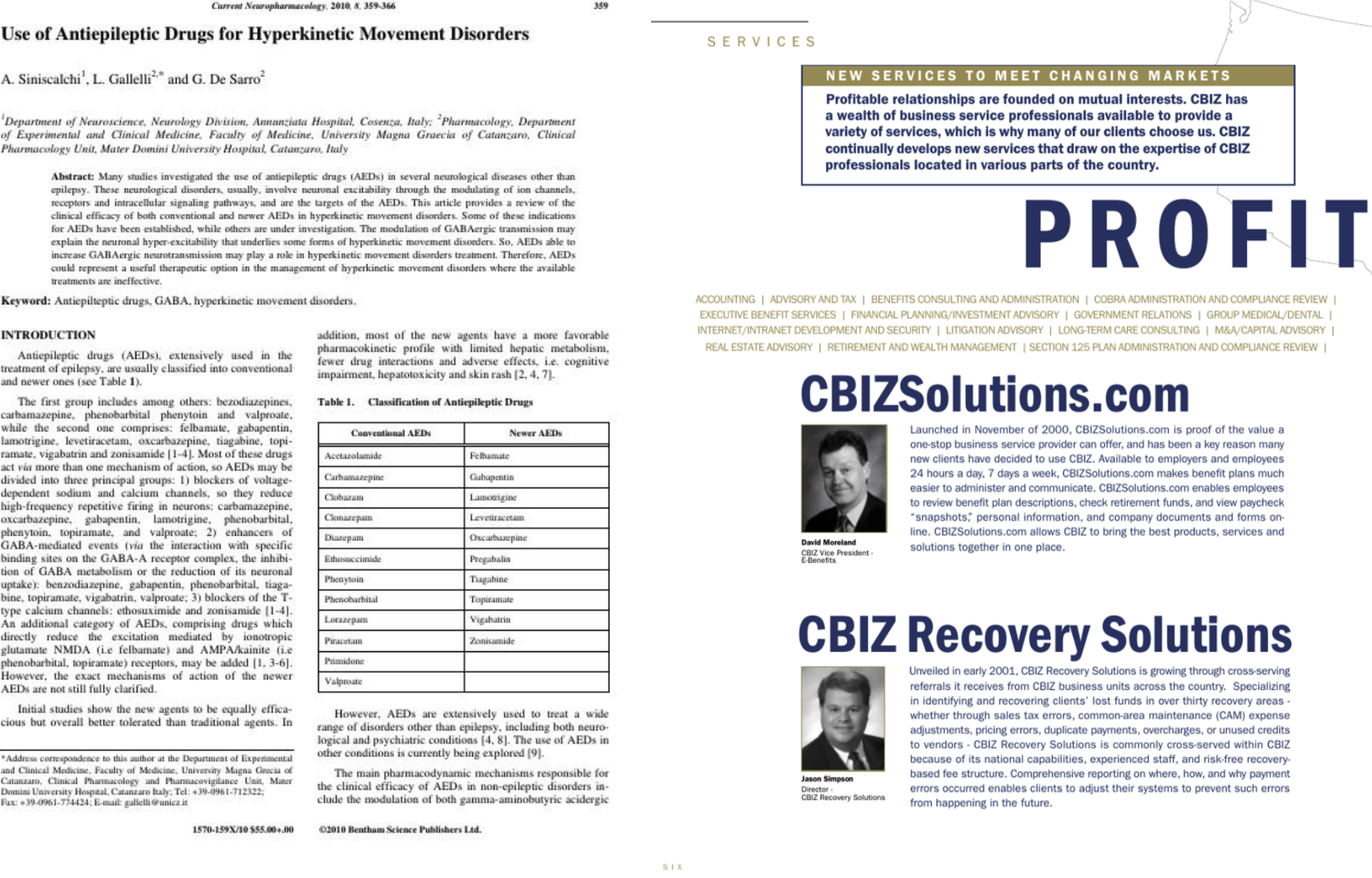}
     \caption{Example document pages from PubLayNet (left) and DocLayNet (right) datasets}
     \label{fig:doc_example}
\end{figure}

PubLayNet \cite{zhong2019publaynet} is a dataset consisting of 358,353 biomedical research articles in PDF format (approximately 335k for training, 11k for development, and 11k for testing). Each document is annotated with bounding boxes and labels for every distinct semantic segment of content (i.e., text, title, list, figure, and table). These bounding boxes are automatically generated by matching the PDF format and the XML format of the articles. In practice, this leads to bounding boxes that are often not perfectly aligned with the underlying document. \cite{li_gat_2017} finds that this process severely degrades the performance of any method that generates boxes using data directly from the underlying document. For an equal comparison across baselines, we maintain these boxes without adjustment and report our raw mAP (mean average precision) score.

In this work, we primarily focused on using the DocLayNet \cite{pfitzmann2022doclaynet} dataset, due to it having a diverse set of unique layouts. While PubLayNet only contains consistently formatted journal articles, DocLayNet spans six distinct genres (i.e., financial reports, manuals, scientific articles, laws \& regulations, patents, and government tenders). It provides layout annotation bounding boxes for 11 different categories (caption, footnote, formula, list-item, page footer, page header, picture, section header, table, text, and title) on 80,863 unique document pages. The documents are divided into a training set (69,375 documents), a development set (6,489 documents), and a testing set (4,999 documents).

Unlike PubLayNet, DocLayNet is a human-annotated dataset; the ground truth bounding boxes are generated by human annotators hand-drawing boxes on top of the documents. These human-drawn boxes then are adjusted to match the underlying extracted text boxes, removing any additional whitespace or annotator errors in the process. This ensures a perfect correlation between the bounding boxes defined by the extracted data and the ground truth labels. For classes that are not well aligned with the underlying text boxes (e.g., picture and table), the boxes are left unmodified. The current state-of-the-art DLA model for this dataset is an off-the-shelf YOLO v5x6 \cite{yolov5} model.

\subsection{Data Pre-processing}

In conventional VRDU-based methods, PDFs are first converted into images before inference. However, many ground truth features --- not only the raw text but text and artifact formatting --- can be lost in the image conversion step if the original file information is discarded. Thus, we parse text boxes directly from the PDF documents, which allows us to acquire precise text content, position, and font information.

When using the DocLayNet dataset, we directly use the provided parsed text boxes, which include the bounding box location, the text within the box, and information about its font. These boxes are only provided for textual content, as opposed to figures, tables, etc. 

When using the PubLayNet dataset, text boxes are parsed directly from original PDF pages using an in-house PDF parser based on \textit{pdfminer.six} \cite{pdfminersix}. To better match DocLayNet, we ignore image-based objects in the document entirely. We apply a series of data cleaning steps (e.g. ignore random symbols in some documents by removing very small bounding boxes smaller than 10 pixels in width and height along with extra white space) on the parsed boxes. To reduce the number of unnecessary edge classifications between adjacent nodes, we use simple distance-based heuristics to merge adjacent boxes. This simplifies the graph structure and reduces the overall memory requirements during training and inference.

Both DocLayNet and PubLayNet contain ground truth data in the COCO format. In order to convert this format to labels on our graph, we take the bounding box for each annotation and find all of the text boxes in the document that have some overlap. We compute the minimum spanning box of this set of text boxes and compare it with the original ground truth. For DocLayNet, if the IoU of the ground truth box and the box formed by all the overlapping text boxes is less than 0.95, we remove text boxes in order of their total area outside the annotation box and select the configuration that achieves the highest IoU.

Once we have associated every annotation with its set of text boxes and their corresponding nodes on the graph, we label each node on the graph with the class of the annotation. Every edge that connects two nodes within the same annotation is labeled a positive edge, while every edge connecting nodes from different annotations is labeled as a negative edge. An example of the labeled graph is presented in \cref{fig:graph_example}.



\begin{table*}[h]
\begin{center}
\caption{Model performance mAP@IoU[0.5:0.95] on the DocLayNet test set, as compared to metrics reported in \cite{pfitzmann2022doclaynet}.}
\label{table:map-doclaynet}
\begin{tabular}{ l | *{5}{X{15mm}} | X{15mm}}
  & MRCNN R50 & MRCNN R101 & FRCNN R101 & YOLO v5x6 & GLAM  & GLAM + YOLO v5x6 \\
\hline
caption & 68.4 & 71.5 & 70.1 & \textbf{77.7} & 74.3  & 77.7 \\
footnote & 70.9 & 71.8 & 73.7 & \textbf{77.2} & 72.0  & 77.2 \\
formula &  60.1 & 63.4 & 63.5 & 66.2  & \textbf{66.6} & 66.6  \\
list item & 81.2 & 80.8 & 81.0 & \textbf{86.2}  & 76.7 & 86.2  \\
page footer & 61.6 & 59.3 & 58.9 & 61.1  & \textbf{86.2} & 86.2  \\
page header & 71.9 & 70.0 & 72.0 & 67.9  & \textbf{78.0}  & 78.0  \\
picture & 71.7 & 72.7 & 72.0 & \textbf{77.1} & 5.7  & 77.1 \\
section header & 67.6 & 69.3 & 68.4 & 74.6 & \textbf{79.8} & 79.8  \\
table & 82.2 & 82.9 & 82.2 & \textbf{86.3} & 56.3  & 86.3\\
text & 84.6 & 85.8 & 85.4 & \textbf{88.1} & 74.3  & 88.1 \\
title & 76.6 & 80.4 & 79.9 & 82.7  & \textbf{85.1} & 85.1  \\
\hline
overall & 72.4 & 73.5 & 73.4 & \textbf{76.8} & 68.6 & \textbf{80.8}
\end{tabular}
\end{center}
\end{table*}


\subsection{Results}

\subsubsection{DocLayNet} As shown in Table~\ref{table:map-doclaynet}, GLAM achieves competitive performance while being over 10$\times$ smaller than the smallest comparable model --- and over 35$\times$ smaller than the current state-of-the-art. Our model outperforms all of the object detection models on five of the 11 total classes, yielding an overall mAP of 68.6. The SOTA model achieves a 76.8. 

\begin{table*}[h]
\begin{center}
\caption{Model performance mAP@IoU[0.5:0.95] on the PubLayNet val set.}
\label{table:map-publaynet}
\begin{tabular}{  l | *{3}{X{15mm}} | *{2}{X{15mm}} } 
 & LayoutLM v3 \cite{huang2022layoutlmv3} & MBC \cite{li_gat_2017}  & VSR \cite{10.1007/978-3-030-86549-8_8} & YOLO v5x6   & GLAM \\
\hline
text & 94.5 & 88.8	& \textbf{96.7} & 91.7 & 87.8 \\
title & 90.6 & 52.8 & \textbf{93.1} & 79.1 & 80.0 \\
list item & \textbf{95.5} & 88.1 & 94.7 & 86.9 & 86.2 \\
table & \textbf{97.9} & 97.7 & 97.4 & 96.5 & 86.8 \\
figure & \textbf{97.0} & 84.0 & 96.4 & 94.7 & 20.6 \\
\hline
overall & \textbf{95.1} & 82.3 &  \textbf{95.7} &  89.8 & 72.2 
\end{tabular}
\end{center}
\end{table*}

\begin{figure}[ht!]
  \centering
    \includegraphics[width=4 in]{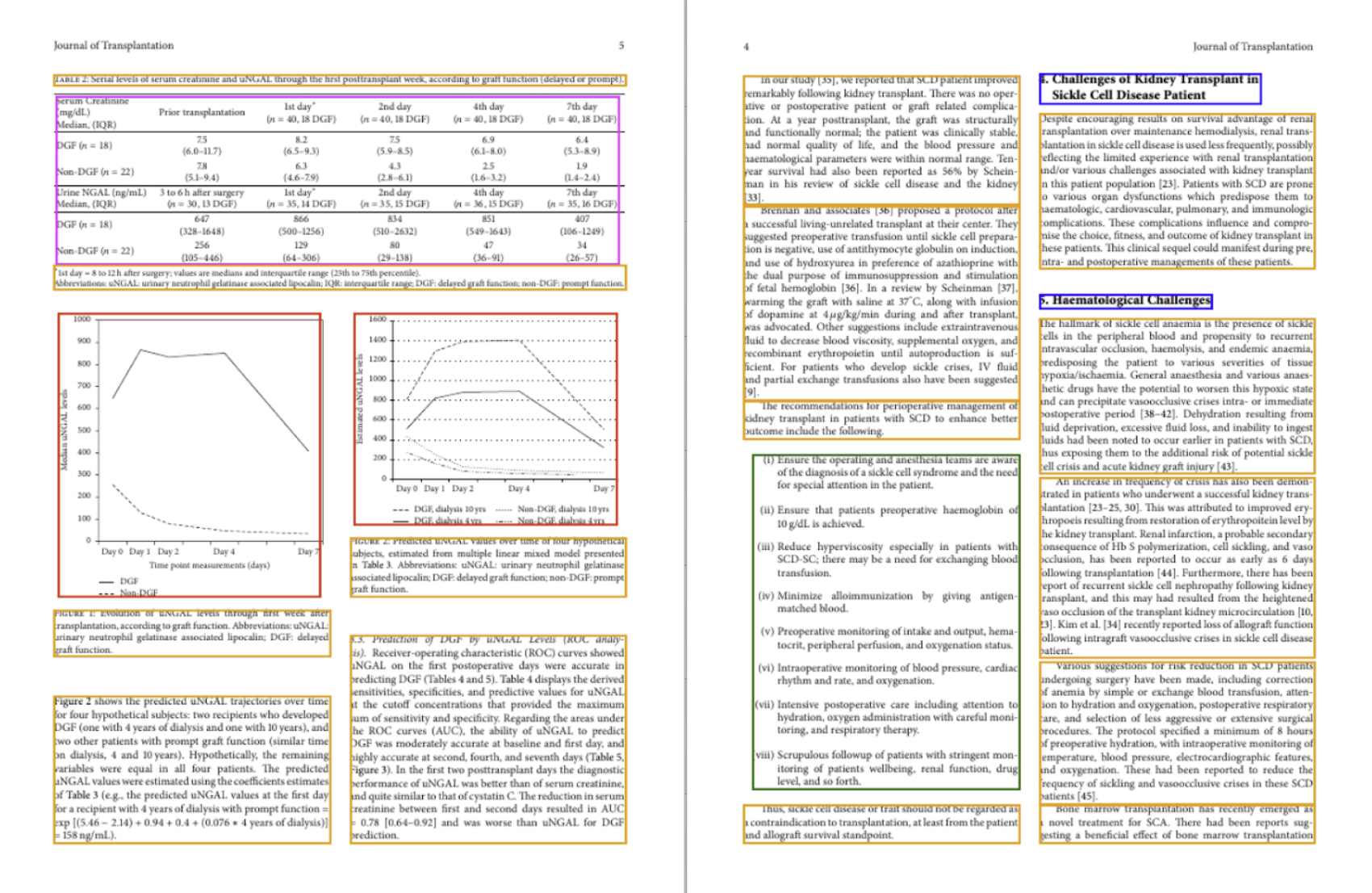}
    \caption{Examples of GLAM predictions on PubLayNet documents. Bounding boxes in orange, blue, green, magenta, and red colors represent predicted segments in text, title, list, table, and figure categories, respectively. Due to the inability to parse images from PDF, the GLAM model underperforms in figure categories.}
    \label{fig:gnn-model-pred}
\end{figure}

Nine of the 11 classes have annotations that involve text box snapping; the tables and pictures classes do not. When considering only these nine classes, our model achieves an mAP score of 77.0, compared to 75.7 for the next-best model. As expected, our model does not perform as well for tables, and it is particularly low-performing for pictures. When only considering the parsed text of tables (and thus disregarding visual markers such as table borders), our model yields a low mAP for the table class. In fact, GLAM performs well in extracting the compact regions of text from table contents (Fig.~\ref{fig:gnn-model-pred}). If we consider all 10 text-based classes (including tables but excluding pictures), GLAM achieves an mAP score of 74.9, compared to 76.8 from the SOTA model.
By creating a simple ensemble with GLAM and YOLO v5x6, we are able to increase the state of the art from 76.8 to 80.8.

\subsubsection{PubLayNet} 

\begin{figure}[ht!]
  \centering
    \includegraphics[width=2.5 in]{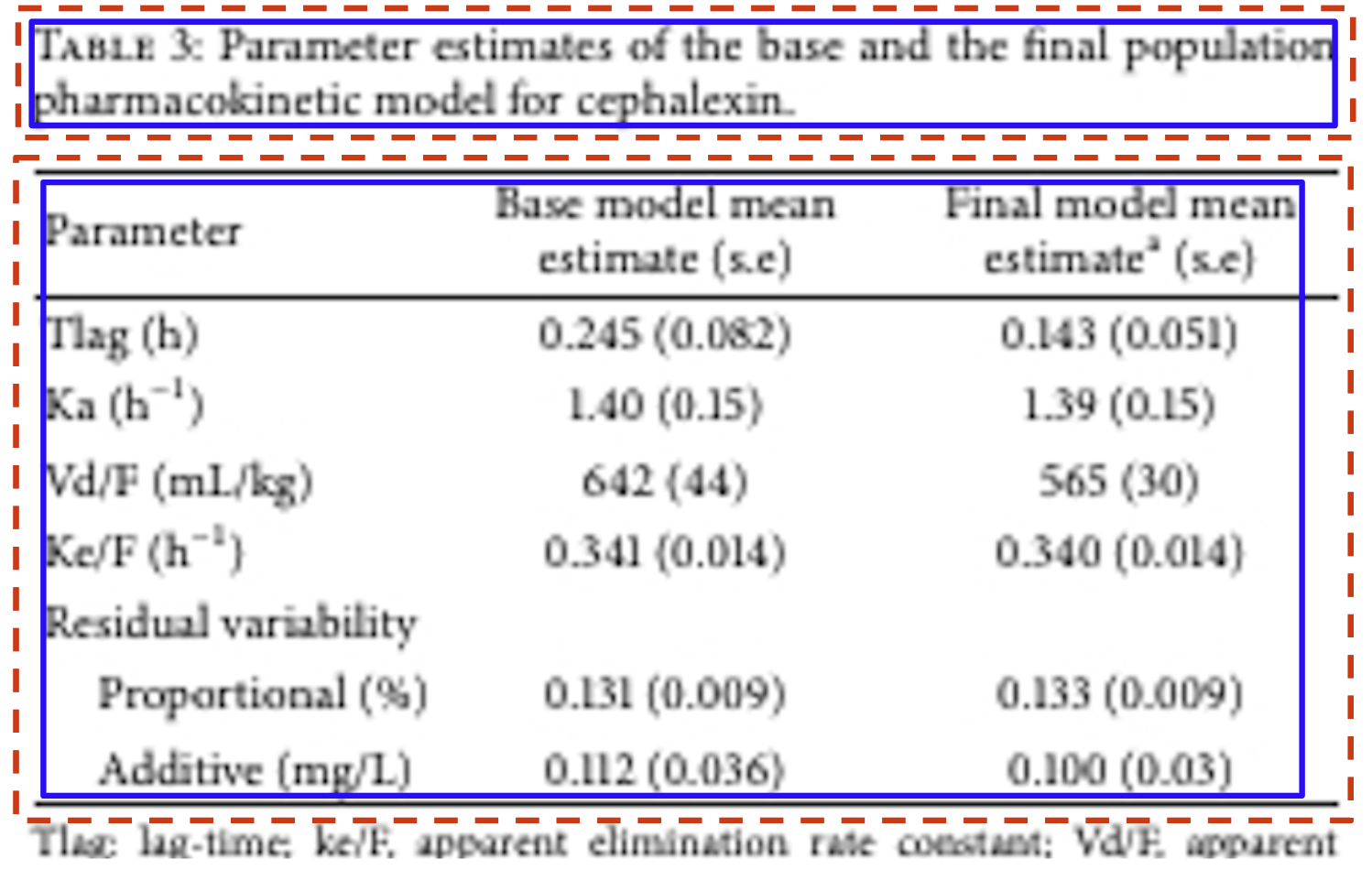}
    \caption{Examples of bounding box misalignment between GLAM prediction and annotation in PubLayNet. Red bounding boxes are from annotation, and blue bounding boxes are from GLAM prediction. GLAM predictions have tight bounding boxes around text-based segments.}
    \label{fig:publaynet-pred}
\end{figure}

The results on PubLayNet are shown in Table~\ref{table:map-publaynet}. On this dataset, our model is significantly handicapped by the misalignment between the text boxes and the ground truth boxes as shown in Figure~\ref{fig:publaynet-pred}. Since it cannot improve the alignment of the original text boxes, this enacts a hard upper bound on performance. The model can only achieve an IoU as high as the overlap between the ground-truth label and the underlying PDF data. Although our method may correctly identify and label all of the underlying nodes, it may still not achieve an IoU high enough to register as a correct classification at most of the thresholds.

We can get a more accurate picture of performance by looking at the lower IoU thresholds. At a threshold of 0.5, GLAM achieves an mAP higher than 90 for all of the text classes. For the ``title'' class, the gap between mAP at this threshold and the final value averaged over the full threshold range is over 17\%. As a comparison, this gap on DocLayNet is less than 7\% for every class. Even with this misalignment, the GLAM model still is able to achieve 88\% of the performance of the SOTA model on the text-based categories.

\subsubsection{Analysis} 
In general, our model performs strongest on small, text-box-based classes, such as titles and headers. We hypothesize that this is mostly due to its ability to perfectly match its bounding box to the ground truth label. Since GLAM has access to the exact text boxes in question, it can output pixel-perfect annotations by simply segmenting the relevant boxes. Object detection models on the other hand are known to struggle with smaller objects \cite{smallod}. 

Conversely, our model performs worst on classes that are not well captured by the underlying text boxes in the document. This is a fundamental limitation of our structural representation of the document. This discussion is continued in Section \ref{sec:limits}.

\subsection{CV Feature Ablation}


\begin{table*}[h]
\centering
\begin{center}
\caption{CV Feature extraction Ablation, mAP@IoU[0.5:0.95] on DocLayNet test set.}
\label{table:ab-doclaynet}
\begin{tabular}{ l | c c | c } 
& GLAM  & GLAM-CV   & $\Delta$ \\
\hline
caption & 74.3  & 60.6  & 13.7 \\
footnote & 72.0  & 61.1  & 10.9 \\
formula & 66.6 & 43.8  & 22.8 \\
list item & 76.7 & 61.5  & 15.2 \\
page footer & 86.2 & 87.3  & -1.1 \\
page header & 78.0  & 78.1 & -0.1 \\
picture & 5.7   & 4.3   & 1.4 \\
section header & 79.8 & 70.1  & 9.7 \\
table & 56.3  & 50.2  & 6.1 \\
text & 74.3  & 68.4  & 5.9 \\
title & 85.1 & 72.9  & 12.2 \\
\hline
overall & 68.6 & 59.9 & 8.79 \\
\end{tabular}
\end{center}
\end{table*}

\begin{table*}[h]
\begin{center}
\caption{CV Feature extraction Ablation, mAP@IoU[0.5:0.95] on PubLayNet val set.}
\label{table:ab-publaynet}
\begin{tabular}{  l | c c | c  } 
& GLAM & GLAM - CV & $\Delta$ \\
 \hline
text & 87.8 & 85.9 & 1.9 \\
title & 80.0 & 79.8 & 0.2 \\
list item & 86.2 & 84.8 & 1.4 \\
table & 86.8 & 86.1 & 0.7 \\
figure & 20.6 & 18.3 & 2.3 \\
\hline
overall & 72.2 & 71.0 & 1.2 \\
\end{tabular}
\end{center}
\end{table*}

We include an ablation study of the computer vision feature extractor in Tables~\ref{table:ab-doclaynet} and~\ref{table:ab-publaynet}. For these models, we only train the graph network component of GLAM, without the CV features. The GNN component represents only one-fourth of the total parameters of the GLAM. Removing the vision feature extraction leads to a consistent decrease in performance across both tasks, albeit only slightly on PubLayNet. This is likely the result of two factors: (1) our model nearing the performance limit imposed by the misaligned boxes in PubLayNet; and (2) the visual complexity of the DocLayNet data. DocLayNet layouts contain many visual cues that are not captured by the underlying text boxes (e.g., bullet points, lines, and colored backgrounds). Results on the PubLayNet dataset (Table~\ref{table:ab-publaynet}) indicate that the GNN component in GLAM already performs well on the visual-simple type of documents like research articles.

\subsection{Model Size and Efficiency}

\begin{table}[h]
\begin{center}
\caption{Comparison of model size and efficiency. All numbers are bench-marked on a G4 AWS instance, with NVIDIA Tesla T4 GPU.}
\label{table:efficiency}
\footnotesize
\begin{tabular}{c | c | c c | c c } 
 \hline
 & Model Size & \multicolumn{2}{c|}{Inference Time (ms)} & \multicolumn{2}{c}{Pages/sec}  \\
& (M) & GPU &  CPU & GPU  & CPU \\

 \hline
LayoutLMv3 & 133   & 687   & 6100 & 1.5 & 0.2 \\
YOLO v5 & 140.7    & 56    & 179  & 17.8 & 5.6\\
\hline
GLAM & 4         & 10   & 16    & 98.0 & 63.7\\
GLAM - CV & 1    & 4    & 9     & 243.9 & 109.9\\
 \hline 
\end{tabular}
\end{center}
\end{table}

Although large, pre-trained models often lead to high performance when adapted to downstream tasks, they tend to be relatively large and slow in training and inference. Both the state-of-the-art object detection models on PubLayNet and DocLayNet have over 130 million parameters. To highlight the efficiency of our graph-based model, in Table~\ref{table:efficiency} we include additional benchmarks regarding inference speed and throughput. We baseline our model against the two highest-performing, open-source models on each task. 

In general, GLAM is much more efficient than the baseline models. The full GLAM model, which achieves competitive performance to the baseline models, is able to process a document page in 10 ms on a single GPU\footnote{We use an NVIDIA Tesla T4 as GPU in our experiments.}, compared to 56 ms and 687 ms for YOLO v5x6 and LayoutLMv3, respectively. Removing the CV feature extraction drops the inference time to 4 ms, allowing the model to process over 243 document pages per second on a single GPU. Such efficient DLA models significantly reduce the engineering complexity in real-world use cases and enable the possibility of on-device model deployments.

\section{Limitations}
\label{sec:limits}
The main limitation of GLAM is that it is highly dependent on the parsing quality of the document. The benefit of using off-the-shelf object detection models is that their performance is much less sensitive to the structure of the underlying document. That is, they can account for the misaligned ground-truth boxes, such as the ones in PubLayNet. Importantly, they can also properly work with scanned documents and other documents with little or no structure -- alleviating the need to address pre-processing decisions. For the same reason, GLAM is unable to capture many regions of vision-rich segments (e.g., figures and pictures) in PDF documents. GLAM also does not take into account any of the available document semantic information. To address this limitation, we could use an additional text feature extractor (text embedding) for the text boxes. 

\section{Conclusion}
In this work, we introduce GLAM, a lightweight graph-network-based method for performing document layout analysis on PDF documents. We build upon a novel, structured representation of the document created from directly parsing the PDF data. We frame the document layout analysis task as performing node and edge classification on document graphs. With these insights, GLAM performs comparably to competitive baseline models while being orders of magnitude smaller. The model works particularly well for text-based layout elements such as headers and titles. By ensembling this model with an off-the-shelf object detection model, we are able to set state-of-the-art performance in DocLayNet. 
We demonstrate that GLAM is a highly-efficient and effective approach to extracting and classifying text from PDF documents, which makes it competitive, especially in real-world applications with limited computational resources.

\bibliographystyle{splncs04}
\bibliography{paper}

\end{document}